\documentclass[letterpaper]{article} 
\usepackage{aaai2026}  
\usepackage{times}  
\usepackage{helvet}  
\usepackage{courier}  
\usepackage[hyphens]{url}  
\usepackage{graphicx} 
\usepackage{natbib}  
\usepackage{caption} 
\usepackage{algorithm}
\usepackage{algorithmic}
\usepackage{url}            
\usepackage{booktabs}       
\usepackage{amsfonts}       
\usepackage{nicefrac}       
\usepackage{microtype}      
\usepackage{xcolor}         
 \usepackage{amsmath}
\usepackage{multirow}
\usepackage{graphicx}
\usepackage{amsthm}  
\usepackage{subfigure}

\newtheorem{theorem}{Theorem}

\usepackage{newfloat}
\usepackage{listings}
\DeclareCaptionStyle{ruled}{labelfont=normalfont,labelsep=colon,strut=off} 
\floatstyle{ruled}
\newfloat{listing}{tb}{lst}{}
\floatname{listing}{Listing}
\title{RankList – A Listwise Preference Learning Framework for \\ Predicting Subjective Preferences}
\author{
    Abinay Reddy Naini\textsuperscript{\rm 1,\rm 2},
    Fernando Diaz\textsuperscript{\rm 1},
    Carlos Busso\textsuperscript{\rm 1}
}
\affiliations{
    \textsuperscript{\rm 1}Language Technologies Institute (LTI), Carnegie Mellon University, USA\\
    \textsuperscript{\rm 2}The University of Texas at Dallas, USA
}

\usepackage{bibentry}
\nocopyright
\begin{document}

\maketitle

\begin{abstract}
Preference learning has gained significant attention in tasks involving subjective human judgments, such as \emph{speech emotion recognition} (SER) and image aesthetic assessment. While pairwise frameworks such as RankNet offer robust modeling of relative preferences, they are inherently limited to local comparisons and struggle to capture global ranking consistency. To address these limitations, we propose RankList, a novel listwise preference learning framework that generalizes RankNet to structured list-level supervision. Our formulation explicitly models local and non-local ranking constraints within a probabilistic framework. The paper introduces a log-sum-exp approximation to improve training efficiency. We further extend RankList with skip-wise comparisons, enabling progressive exposure to complex list structures and enhancing global ranking fidelity. Extensive experiments demonstrate the superiority of our method across diverse modalities. On benchmark SER datasets (MSP-Podcast, IEMOCAP, BIIC Podcast), RankList achieves consistent improvements in Kendall's Tau and ranking accuracy compared to standard listwise baselines. We also validate our approach on aesthetic image ranking using the Artistic Image Aesthetics dataset, highlighting its broad applicability. Through ablation and cross-domain studies, we show that RankList not only improves in-domain ranking but also generalizes better across datasets. Our framework offers a unified, extensible approach for modeling ordered preferences in subjective learning scenarios.
\end{abstract}

\section{Introduction}
Ranking is a central problem in machine learning, especially in tasks where the goal is to order items based on subjective or contextual criteria. Unlike classification or regression, where absolute labels are predicted, ranking methods model relative orderings, making them better suited for tasks such as retrieval, recommendation, and subjective assessments \cite{liu2009learning}. This study aims to predict full global rankings, where the relative ordering of all items is of interest. 

Global optimization contrasts with many existing methods in several ways.  Traditional \emph{learning-to-rank} (LTR) methods are generally categorized as pointwise, pairwise, or listwise. Pointwise methods treat ranking as independent regression or classification problems \cite{li2007mcrank}, which often ignore inter-item dependencies and fail to model ordinal structure effectively. Pairwise approaches such as RankNet \cite{Burges_2005} attempt to minimize inversions by learning relative preferences between sample pairs, but they inherently focus only on local relationships and can miss global ranking structure \cite{joachims2002optimizing, herbrich2000large}. Although listwise methods aim to overcome these limitations by directly optimizing over entire ranked lists, these methods often focus on objectives that emphasize the accuracy at the top of the ranked list.  Notable examples include ListNet \cite{cao2007learning}, ListMLE \cite{xia2008listwise}, and ApproxNDCG \cite{qin2010general}, which capture inter-item dependencies and often align more closely with ranking evaluation metrics. However, these methods often rely on permutation-based formulations or metric-specific surrogates \cite{cao2007learning,xia2008listwise, qin2008listwise, burges2010ranknet}, which are tightly coupled to retrieval-oriented objectives and may not generalize well to perceptual ranking tasks. For example, in ranking speech samples by valence, accurate ordering across the entire list is essential, not just the top-ranked items. Furthermore, in subjective assessment tasks such as emotion recognition or aesthetic assessment, predicting the relative ranking of samples often provides a more stable and interpretable alternative to regressing on absolute consensus scores, which are known to be noisy and subjective \cite{Yannakakis_2017, Parthasarathy_2017}. Additionally, existing listwise models are not well tailored to applications involving sparse and implicitly ordered preferences. 

To address the challenges of current listwise methods, we introduce a novel listwise preference learning formulation that extends RankNet's pairwise loss into a robust list-level loss. Our proposed loss considers both adjacent and skip-connected pairs within an ordered list, capturing both local and global relational constraints. We introduce a log-sum-exp approximation to enable efficient optimization. The RankList framework, which combines these strategies, improves training stability and listwise generalization without explicit permutation-based modeling. While prior listwise models have been developed for traditional LTR benchmarks (e.g., web search), such datasets emphasize top-heavy metrics and assume consistent rank positions; these settings are not consistent with the needs of perceptual ranking, where global performance is emphasized. Therefore, our experiments investigate RankList on two perceptual ranking tasks: \emph{speech emotion recognition} (SER) and aesthetic image quality assessment. Within corpus and cross-corpus experiments on diverse benchmarks, including the MSP-Podcast \cite{Lotfian_2019_3}, BIIC-Podcast \cite{Upadhyay_2023_2}, IEMOCAP \cite{Busso_2008_5}, and MSP-IMPROV \cite{Busso_2017} databases, demonstrate that RankList consistently outperforms pairwise and listwise baselines, showing superior generalization across SER tasks. RankList produced performance gains of approximately 11\% and 7.5\% relative improvement in Kendall's Tau over RankNet (pairwise) and the best-performing listwise baseline, respectively. The approach is validated in the task of artistic image aesthetics assessment, showing the generalization of the proposed formulation. 

\begin{itemize}
    \item We propose RankList, a simple but effective listwise framework building on the RankNet strategy that integrates adjacent and skip-wise comparisons to model structured ranking constraints.
    
    \item We propose a log-sum-exp approximation to enable efficient and stable training, avoiding the need for permutation-based modeling.
    
    \item RankList achieves SOTA results on SER and aesthetic image ranking, outperforming strong pairwise and listwise baselines while remaining scalable.
\end{itemize}

\section{Background}
Early listwise approaches such as ListNet \cite{cao2007learning} and ListMLE \cite{xia2008listwise} introduced probabilistic modeling over permutations or maximum likelihood over ranked sequences. These formulations provided better alignment with ranking evaluation metrics but introduced significant computational overhead due to their reliance on full list permutations. To address efficiency and stability concerns, later methods such as SoftRank \cite{Taylor_2008}, ApproxNDCG \cite{qin2010general}, and LambdaMART \cite{wu2010adapting} introduced surrogate losses or gradient approximations tailored to optimize specific metrics such as \emph{normalized discounted cumulative gain} (NDCG), and \emph{mean average precision} (MAP). PiRank \cite{swezey2021pirank} introduced differentiable approximations to sorting operators to enable end-to-end training with ranking losses. Such metric-driven losses are optimized for retrieval-focused criteria that prioritize top-ranked accuracy, making them less effective for tasks demanding ordinal consistency over the entire list.

To move beyond these metric-bound constraints, recent work has explored alternative formulations such as neighborhood-based contrastive learning. Notably, \emph{Rank-n-Contrast} (RnC) \cite{zha2023rank} proposes learning smooth representations for regression and ranking by constructing local ranking neighborhoods and using contrastive losses to preserve relative order. While RnC is not a traditional LTR method and does not optimize a list-level ranking loss directly, it demonstrates how leveraging neighborhood structure can be beneficial for learning in low-resource or ambiguous domains. However, it still lacks an explicit listwise formulation and does not leverage sequential ordering constraints critical for structured ranking tasks.

\section{Methodology}
\label{sec:methodology}

In this section, we present a detailed formulation of our proposed listwise preference learning framework, building explicitly on the RankNet pairwise preference learning model. The primary goal is to extend RankNet's pairwise loss formulation to a comprehensive listwise setting, capturing both local and global ranking relationships.

\subsection{Problem Setup and Notation}

Let us consider a set of $N$ samples represented by their corresponding feature vectors $(\Phi_1, \Phi_2, \dots, \Phi_N)$, which are arranged in descending order according to an attribute given by the ground-truth ranking. Our objective is to learn a continuous scoring function: $f: \mathbb{R}^d \rightarrow \mathbb{R}$ that maps each feature vector $\Phi_i \in \mathbb{R}^d$ into a scalar score $s_i = f(\Phi_i)$ such that the relative ordering of these scores accurately reflects the ground-truth ranking of the samples.

\subsection{Cost Function for Pairwise RankNet}
\label{ssec:ranknet}

RankNet \cite{Burges_2005} uses a probabilistic framework to model the relative preference between two samples. The labels correspond to relative preference between pairs of samples (i.e., \emph{sample $i$} is preferred over \emph{sample $j$} for a given attribute). Given a pair of samples $(x_i, x_j)$ with corresponding feature vectors $\Phi_i$ and $\Phi_j$, their respective preference scores are computed as $s_i = f(\Phi_i)$ and $s_j = f(\Phi_j)$. The probability of preferring sample $x_i$ over sample $x_j$ is modeled using a sigmoid function:

\begin{equation}
P_{ij} = \frac{1}{1 + e^{-\sigma(s_i - s_j)}},
\end{equation}

\noindent
where $\sigma$ is a scaling parameter controlling the steepness of the sigmoid function. During training, RankNet minimizes the cross-entropy loss between the predicted preferences $P_{ij}$ and ground-truth binary preferences $\bar{P}_{ij}$, defined as:

\begin{equation}
\bar{P}_{ij} = \begin{cases}
1 & \text{if $x_i$ is preferred over $x_j$,} \\
0 & \text{otherwise.}
\end{cases}
\end{equation}

The cross-entropy loss $\mathcal{C_R}$ for a pair of samples $(x_i, x_j)$ is:

\begin{align}
\mathcal{C_R} &= -\bar{P}_{ij} \log P_{ij} - (1 - \bar{P}_{ij}) \log(1 - P_{ij}) \\
&= \begin{cases}
\log\left(1 + e^{-\sigma(s_i - s_j)}\right) & \text{if } \bar{P}_{ij} = 1, \\
\log\left(1 + e^{-\sigma(s_j - s_i)}\right) & \text{if } \bar{P}_{ij} = 0.
\end{cases}
\end{align}

\subsection{Proposed RankList Preference Learning Cost}
\label{ssec:listwise}

This study extends RankNet to a listwise scenario. We consider an ordered list of $N$ random samples (i.e., $s_1>s2>\ldots > s_N$) and their corresponding preference scores:

\begin{equation}
s_1 = f(\Phi_1), \quad s_2 = f(\Phi_2), \quad \dots, \quad s_N = f(\Phi_N).
\end{equation}

To explicitly model local ordering constraints, we define pairwise differences between adjacent scores as:

\begin{equation}
O_{i(i+1)} = s_i - s_{i+1}, \quad \text{for } i \in \{1, 2, \dots, N-1\}.
\end{equation}

Applying RankNet's cost function to these adjacent pairs, the local pairwise loss for each adjacent pair $(i, i+1)$ is defined as:

\begin{equation}
\mathcal{L}_i = \log\left(1 + e^{-\sigma O_{i(i+1)}}\right).
\end{equation}

Aggregating these local pairwise losses, the total listwise preference learning cost is given by:

\begin{equation}
\mathcal{L}_{\text{listwise}} = \sum_{i=1}^{N-1} \mathcal{L}_i = \sum_{i=1}^{N-1} \log\left(1 + e^{-\sigma(s_i - s_{i+1})}\right).
\end{equation}

This cost function only considers pairwise relations between adjacent samples in the list. To improve the model's capability of capturing global ranking relationships beyond adjacent pairs within the selected list, we extend the above formulation by incorporating skip-term comparisons between non-adjacent samples. These additional comparisons capture broader contextual information within each list. Specifically, we define skip-$k$ term differences:

\begin{equation}
O_{i(i+k)} = s_i - s_{i+k+1}, \quad \text{for } k \geq 1 \text{ and } i \leq N - k-1.
\end{equation}

This formulation leads to a generalized listwise loss function that includes both adjacent and non-adjacent (skip) term penalties:

\begin{equation}
\label{eq:extended_listwise}
\mathcal{L}_{\text{extended}} = \sum_{k=0}^{K} \sum_{i=1}^{N-k-1} \log\left(1 + e^{-\sigma O_{i(i+k+1)}}\right),
\end{equation}

\noindent
where $K$ is a hyperparameter determining the maximum number of skip levels included (e.g., $K=2$ includes both skip-1 and skip-2 comparisons). The above equation can equivalently be expressed in product form as:

\begin{equation}
\label{eq:product_form}
\mathcal{L}_{\text{extended}} = \log \left( \prod_{k=0}^{K} \prod_{i=1}^{N-k-1} \left(1 + e^{-\sigma O_{i(i+k+1)}}\right) \right).
\end{equation}

While the gradient of \( \mathcal{L}_{\text{extended}} \) is Lipschitz continuous, since it is a finite sum of smooth functions such as \( \log(1 + e^{-z}) \), its Lipschitz constant is unbounded with respect to both the number of skip levels \( K \) and the list length \( N \). Specifically, the gradient magnitude can grow linearly with \( N \times K \), leading to poor conditioning. The additive structure of gradients in \( \mathcal{L}_{\text{extended}} \) causes local noise or misordering to directly affect the magnitude and direction of the overall gradient.

To mitigate this issue, we approximate the sum of logarithmic losses with a log-sum-exp formulation, which performs a soft aggregation of pairwise ranking terms. This simplification allows us to balance the contributions of each term in a numerically stable way, while bounding the gradient magnitude. The extended sum can be written as:

\begin{equation}
\begin{aligned}
\mathcal{L}_{\text{extended}} = \log\Big[ &(1 + e^{-\sigma O_{12}})(1 + e^{-\sigma O_{23}})(1 + e^{-\sigma O_{34}}) \dots \\
&(1 + e^{-\sigma O_{13}})(1 + e^{-\sigma O_{24}})(1 + e^{-\sigma O_{35}}) \dots \Big]
\end{aligned}
\end{equation}

Expanding the above product inside the logarithm yields a series of additive terms due to multiplication:

\begin{equation}
\begin{aligned}
\mathcal{L}_{\text{extended}} = \log\Big[ 
&1 + \sum_{k=0}^{K} \sum_{i=1}^{N-k-1} e^{-\sigma O_{i(i+k+1)}} \\
&+ \sum_{\substack{(i,j) \\ (k_1,k_2)}} e^{-\sigma(O_{i(i+k_1+1)} + O_{j(j+k_2+1)})} + \cdots \Big]
\end{aligned}
\end{equation}

To simplify computation and gradient analysis, we approximate the product by retaining only the first-order additive terms (i.e., the individual exponentials) and discarding higher-order interactions between comparisons (see supplemental material for a discussion on the implications of this truncation). This simplification leads to our final approximation:

\begin{equation}
\label{cost_appr}
\mathcal{L}_{\text{RankList}} = \log\left(1 + \sum_{k=0}^{K} \sum_{i=1}^{N-k-1} e^{-\sigma O_{i(i+k+1)}}\right)
\end{equation}

By incorporating both adjacent and skip-term comparisons within a unified log-sum-exp framework, \(\mathcal{L}_{\text{RankList}}\) offers bounded and smooth optimization dynamics. It preserves ranking fidelity through softly normalized pairwise interactions, avoiding unbounded gradient growth and instability.

An alternative approach could involve normalizing \( \mathcal{L}_{\text{extended}} \) with a factor such as \( \frac{1}{NK} \) to control gradient scale. However, this uniform averaging treats all comparisons equally, which also dilutes the influence of more informative gradients during training. In contrast, the log-sum-exp approximation acts as a soft surrogate of the max, enabling softmax-like prioritization of difficult (misordered) comparisons. This formulation helps preserve stronger and more stable gradient flow, whereas uniform averaging can lead to vanishing gradients due to the contribution of many small near-zero terms.

\begin{theorem}[Smoothness and Bounded Gradient of \(\mathcal{L}_{\text{RankList}}\)]
\label{theorem:approx_stability}
The approximate listwise loss \(\mathcal{L}_{\text{RankList}}\), defined in Equation \ref{cost_appr} is Lipschitz continuous, and its gradient is also Lipschitz continuous. Furthermore, the gradient norm is uniformly bounded, independent of the list size \(N\) and skip parameter \(K\), unlike the extended loss formulation.
\end{theorem}

\begin{proof}
Let us denote
\[
Z = \sum_{k=0}^{K} \sum_{i=1}^{N-k-1} e^{-\sigma O_{i(i+k+1)}}, \quad \text{so that} \quad \mathcal{L}_{\text{RankList}} = \log(1 + Z).
\]
Applying the chain rule, the gradient becomes:
\[
\nabla \mathcal{L}_{\text{RankList}} = \frac{1}{1 + Z} \cdot \nabla Z.
\]
For any score component \(s_j\), the derivative of \(Z\) is given by:
\[
\frac{\partial Z}{\partial s_j} = \sum_{k=0}^{K} \sum_{i=1}^{N-k-1} \pm \sigma e^{-\sigma O_{i(i+k+1)}} \cdot \mathbb{I}\{j = i \text{ or } j = i+k+1\},
\]
where the sign depends on the role of \(s_j\) in the comparison. Since each term is non-negative and \(|\pm \sigma e^{-\sigma O}|\leq \sigma\), we have:
\[
\|\nabla Z\| \leq \sigma Z \quad \Rightarrow \quad \|\nabla \mathcal{L}_{\text{RankList}}\| \leq \frac{\sigma Z}{1 + Z} \leq \sigma.
\]
Hence, the gradient norm is globally bounded by \(\sigma\), and the loss is Lipschitz continuous with constant at most \(\sigma\), independent of \(N\) or \(K\).

In contrast, for the original extended formulation:
\[
\mathcal{L}_{\text{extended}} = \sum_{k=0}^{K} \sum_{i=1}^{N-k-1} \log(1 + e^{-\sigma O_{i(i+k+1)}}),
\]
each term contributes an individual gradient \(\frac{1}{1 + e^{\sigma O_{i(i+k+1)}}} \leq 1\), leading to a cumulative gradient norm:
\[
\|\nabla \mathcal{L}_{\text{extended}}\| \leq C_1 NK,
\]

\noindent
where \(C_1\) is the bound from individual terms. This result implies that as \(N\) or \(K\) increases, the gradient magnitude can grow unbounded, potentially causing instability in training. Thus, \(\mathcal{L}_{\text{RankList}}\) provides better gradient control and smoother optimization dynamics.
\end{proof}

\subsection{Approximation Issues and Mitigation Strategy}

The log-sum-exp approximation significantly reduces computational complexity by avoiding explicit summation over multiple log terms. While this truncation removes higher-order interactions, it keeps the most important misordered terms and provides a smoother approximation that focuses on the most meaningful ranking errors. However, this approximation can lead to numerical instability when the score differences $O_{i(i+k)}$ are small, resulting in exponential terms that dominate and destabilize gradients during training. To counteract this problem, we adopt a robust pre-training strategy wherein the scoring function $f(\cdot)$ is first trained using the original RankNet pairwise loss. This initial phase ensures that the model learns stable and sufficiently separated score differences. Once a stable initialization is achieved, we fine-tune the model using the listwise approximation loss $\mathcal{L}_{\text{RankList}}$ described above. This curriculum-style training helps avoid the undesirable behavior of the exponential terms early in optimization and leads to more consistent convergence across datasets and tasks.

\section{Applications: Preference Learning for Speech Emotion Recognition}
\label{sec:ser_application}

We evaluate the effectiveness of the proposed RankList framework on tasks involving subjective human judgments. This Section presents our evaluation on \emph{speech emotion recognition} (SER). This domain is especially well-suited for listwise preference learning due to its dependence on perceptual annotations that produce absolute scores with a high level of disagreement across evaluators \cite{Yannakakis_2017}. The challenge lies in modeling subtle ordinal relationships from noisy subjective ratings. Modeling ordinal SER formulation is an active research area \cite{Cao_2015, Lei_2023,Lotfian_2016_2, Lotfian_2016, Naini_2023_3, Naini_2023_2, Naini_2023, Parthasarathy_2017}, leveraging the ordinal nature of emotions \cite{Yannakakis_2017, Yannakakis_2021}.  

\subsection{Task Overview and Data Setup}

We aim to learn a model that ranks speech segments based on emotional attributes such as arousal (calm to active), valence (negative to positive), and dominance (weak to strong). We use the MSP-Podcast corpus \cite{Lotfian_2019_3}, a large, naturalistic dataset annotated with continuous emotional scores. These annotations reflect perceived intensity levels along the three core emotional dimensions and are provided as real-valued scores in the range [1, 7].

To create relative preference labels suitable for training, we construct lists of $N$ speech segments ordered by their annotated attribute scores. A minimum score margin $W$ is enforced between any two samples within a list to ensure clear relative preferences and reduce label noise. Formally, for any selected list:

\begin{equation}
\min_{i, j \in \{1, 2, \dots, N\}, i \neq j} |\text{score}(x_i) - \text{score}(x_j)| \geq W.
\label{eq:margin}
\end{equation}

A larger $W$ ensures higher label confidence but reduces the number of available training lists. During testing, we repeatedly extract subsets of evaluation with 200 samples and compare their predicted rankings against the ground-truth rankings to compute the \emph{Kendall's Tau} (KT) metric and pairwise accuracy (acc). These metrics are averaged across all subsets to obtain a robust estimate of model performance, capturing generalization to both subtle and prominent emotional differences. To determine if the improvements are statistically significant, we conduct one-tailed t-tests (significance threshold of $p < 0.05$) for each baseline comparison and apply a Bonferroni correction to account for multiple comparisons. Values marked with $^{\ast}$ indicate that the proposed RankList approach is significantly better than all other baselines.

\subsection{Feature Representation and Model Setup}

To represent the speech segments, we adopt a pre-trained WavLM-Large model \cite{Liu_2021_3} as a front-end feature extractor. We fine-tune WavLM on an emotional attribute prediction task using the annotated scores from the training portion of the MSP-Podcast corpus. The final layer is removed, and the learned representations are then used as fixed input features for all ranking models, including RankList and its baselines. This fixed two-stage setup allows us to isolate the ranking performance of the preference learning methods without confounding factors from raw feature extraction.

We compare RankList against established preference learning baselines, including pairwise RankNet~\cite{Burges_2005}, ListNet~\cite{cao2007learning}, ListMLE~\cite{xia2008listwise}, SoftRank~\cite{Taylor_2008}, and Rank-n-Contrast (RnC) \cite{zha2023rank}. All listwise methods were implemented from scratch for consistency.
Detailed descriptions of the datasets, baselines, training procedures, and hyperparameter settings are provided in the additional material for reproducibility and clarity. For RankList, we report results using the best-performing configuration on the development set. The best version incorporates skip-2 comparisons and is based on the proposed log-sum-exp approximation formulation given in Equation~\ref{cost_appr}.

\subsection{Experimental Results}

Table~\ref{tab:results} presents the performance comparison across multiple preference learning methods on the MSP-Podcast dataset for arousal, valence, and dominance. Among the traditional baselines, ListMLE and SoftRank offer modest gains over the pairwise RankNet framework, reflecting the potential benefit of listwise modeling. Notably, Rank-n-Contrast (RnC)~\cite{zha2023rank}
demonstrates strong performance across all metrics by leveraging neighborhood-level contrastive learning. It achieves the highest performance among all baselines, particularly in terms of accuracy, illustrating the advantages of local structure-aware embedding learning.

Our proposed method, RankList, performs best across all emotional attributes, demonstrating consistent KT and pairwise accuracy improvements. RankList shows enhanced generalization over pairwise and listwise baselines by jointly capturing adjacent and non-adjacent pairwise constraints within the ordered lists. Compared to the next-best baseline (RnC), RankList achieves an average relative improvement of approximately 5\% in KT and 3.5\% in accuracy across all attributes. These gains affirm the value of the proposed skip-term augmented listwise loss and its efficient approximation.

\begin{table}[t]
\centering
\caption{Performance comparison (Kendall's Tau (KT) and accuracy (acc)) on MSP-Podcast for arousal, valence, and dominance.}
\small
\setlength{\tabcolsep}{3.8pt}
\begin{tabular}{l|cc|cc|cc}
\hline
\multirow{2}{*}{Method} & \multicolumn{2}{c|}{Arousal} & \multicolumn{2}{c|}{Valence} & \multicolumn{2}{c}{Dominance} \\
 & KT & Acc & KT & Acc & KT & Acc \\
\hline
Pairwise   & 0.526 & 76.4 & 0.511 & 74.8 & 0.419 & 69.5 \\
ListNet    & 0.504 & 74.2 & 0.498 & 71.8 & 0.403 & 68.1 \\
ListMLE    & 0.530 & 77.1 & 0.508 & 75.2 & 0.425 & 69.3 \\
SoftRank   & 0.521 & 75.8 & 0.514 & 74.0 & 0.411 & 70.2 \\
RnC        & 0.541 & 78.7 & 0.528 & 77.2 & 0.435 & 72.3 \\
\textbf{RankList} & \textbf{0.591}$^{\ast}$ & \textbf{82.9}$^{\ast}$ & \textbf{0.565}$^{\ast}$ & \textbf{80.0}$^{\ast}$ & \textbf{0.461}$^{\ast}$ & \textbf{74.5}$^{\ast}$ \\
\hline
\end{tabular}
\vspace{1mm}
\parbox{0.95\linewidth}{\footnotesize $^{\ast}$ indicates statistically significant improvement over all other baselines.}
\label{tab:results}
\end{table}

\subsection{Ablation Study: Extensions and Variants of RankList}

To better understand the impact of various design choices in our proposed RankList framework, we perform an ablation study comparing several model variants:

\begin{itemize}
\item \textbf{RankList}: Our full proposed model incorporating both adjacent and skip-2 comparisons ($O_{i(i+1)}$, $O_{i(i+2)}$, and $O_{i(i+3)}$), optimized using the log-sum-exp approximation described in Eq.~\ref{cost_appr}. This is the final RankList configuration used in the main evaluation results (Table~\ref{tab:results}).

\item \textbf{RankList\textsubscript{WA (without approx) }}: A variant of RankList that replaces the log-sum-exp approximation with the exact summation over individual RankNet-style log loss terms across all adjacent and skip comparisons (see Eq. \ref{eq:extended_listwise}). This model also uses skip-2 comparisons. 

\item \textbf{RankList\textsubscript{skip1}}: This variant includes only adjacent ($O_{i(i+1)}$) and skip-1 ($O_{i(i+2)}$) comparisons, allowing evaluation of performance with limited non-local constraints.

\item \textbf{RankList\textsubscript{skip3}}: This model includes adjacent, skip-1 ($O_{i(i+2)}$), skip-2 ($O_{i(i+3)}$) and skip-3 ($O_{i(i+4)}$) comparisons, to include the impact of longer-range dependencies.

\item \textbf{RankList\textsubscript{no-skip}}: This model uses only adjacent comparisons ($O_{i(i+1)}$), removing all skip-term dependencies. This model quantifies the benefit of non-local structure modeling.

\item \textbf{RankList\textsubscript{w/o-pt}}: This configuration is identical to the full RankList model, but it skips the initial RankNet-based pairwise pretraining and directly optimizes the listwise objective from random initialization.
\end{itemize}

\begin{table}[t]
\centering
\caption{Ablation study of RankList variants on MSP-Podcast. Metrics: Kendall's Tau (KT) and accuracy (acc).}
\small
\setlength{\tabcolsep}{3.8pt}
\begin{tabular}{l|cc|cc|cc}
\hline
\multirow{2}{*}{Method} & \multicolumn{2}{c|}{Arousal} & \multicolumn{2}{c|}{Valence} & \multicolumn{2}{c}{Dominance} \\
 & KT & Acc & KT & Acc & KT & Acc \\
\hline
RankList & 0.591 & 82.9 & 0.565 & 80.0 & 0.461 & 74.5 \\
RankList\textsubscript{WA} & 0.564$^{\dag}$ & 80.7$^{\dag}$ & 0.538$^{\dag}$ & 77.9$^{\dag}$ & 0.430$^{\dag}$ & 72.1$^{\dag}$ \\
RankList\textsubscript{skip1} & 0.586 & 82.5 & 0.559 & 79.3 & 0.450 & 73.8 \\
RankList\textsubscript{skip3} & 0.580$^{\dag}$ & 82.1 & 0.552$^{\dag}$ & 79.0 & 0.451 & 74.0 \\
RankList\textsubscript{no-skip} & 0.571$^{\dag}$ & 81.3$^{\dag}$ & 0.548$^{\dag}$ & 78.6$^{\dag}$ & 0.442$^{\dag}$ & 72.6$^{\dag}$ \\
RankList\textsubscript{w/o-pt} & 0.582$^{\dag}$ & 81.7 & 0.553$^{\dag}$ & 79.2 & 0.439$^{\dag}$ & 72.8$^{\dag}$ \\
\hline
\end{tabular}
\vspace{1mm}
\parbox{0.95\linewidth}{\footnotesize $^{\dag}$ indicates 
that the performance difference with the full model is statistically significant.}
\label{tab:ablation_ranklist}
\end{table}

Table \ref{tab:ablation_ranklist} presents the performance comparison for these variants on the test set of the MSP-Podcast dataset to assess the contribution of each design choice. The best-performing configuration, denoted as \textit{RankList}, incorporates both skip-1 and skip-2 term comparisons along with the log-sum-exp approximation, serving as the reference setup for the main evaluations reported in Table~\ref{tab:results}.

Removing the approximation component (\textit{RankList\textsubscript{without approx}}) leads to consistent degradation in both KT and accuracy across all emotional attributes. This trend indicates that the approximation not only improves computational efficiency but also contributes to optimization stability and better ordinal preservation. Similarly, excluding skip comparisons entirely (\textit{RankList\textsubscript{no-skip}}) results in notable performance drops, underscoring the importance of incorporating non-adjacent constraints to enhance the global ranking structure.

Variants with different skip-term configurations (\textit{skip1} and \textit{skip3}) illustrate the effect of broadening non-adjacent comparisons in the cost function. \textit{RankList\textsubscript{skip1}}, which includes adjacent and skip-1 terms, yields strong performance across all attributes, demonstrating the benefit of incorporating minimal non-local ranking constraints. Extending this further, \textit{RankList\textsubscript{skip3}} includes up to skip-3 comparisons, but shows no consistent improvement over the base model. In some cases, this model with extended connections slightly degrades performance. This trend indicates that while non-local structure is essential, expanding the comparison horizon beyond skip-2 does not yield additional gains and may introduce redundancy or noise in subjective tasks such as emotion recognition. The full RankList model, which includes adjacent through skip-2 terms, emerges as the best-performing configuration, balancing complexity and expressiveness in modeling global ranking structure.

The \textit{RankList\textsubscript{w/o-pt}} variant, which omits the initial pairwise pre-training phase, still achieves strong results, outperforming many prior listwise models and most ablation variants. This result demonstrates that the RankList formulation is inherently robust and does not rely solely on pre-training for good performance. However, the comparison also affirms that pre-training further refines ranking quality, particularly under noisy or sparse supervision settings, as evidenced by a more noticeable drop in dominance prediction, often the most challenging of the three emotional attributes with \emph{self-supervised learning} (SSL) speech representations such as WavLM \cite{Wagner_2023,Mote_2024, Naini_2024}. Overall, these ablations validate that each component of RankList, including skip-term modeling, log-sum-exp approximation, and pre-training, meaningfully contributes to the final performance. Their combined effect leads to consistent gains in both KT and accuracy, reinforcing the value of structured, extensible listwise formulations for subjective ranking tasks.

\begin{figure}[t]
	\centering
	\subfigure[Performance versus training size]
	{
	\includegraphics[width=0.75\columnwidth]{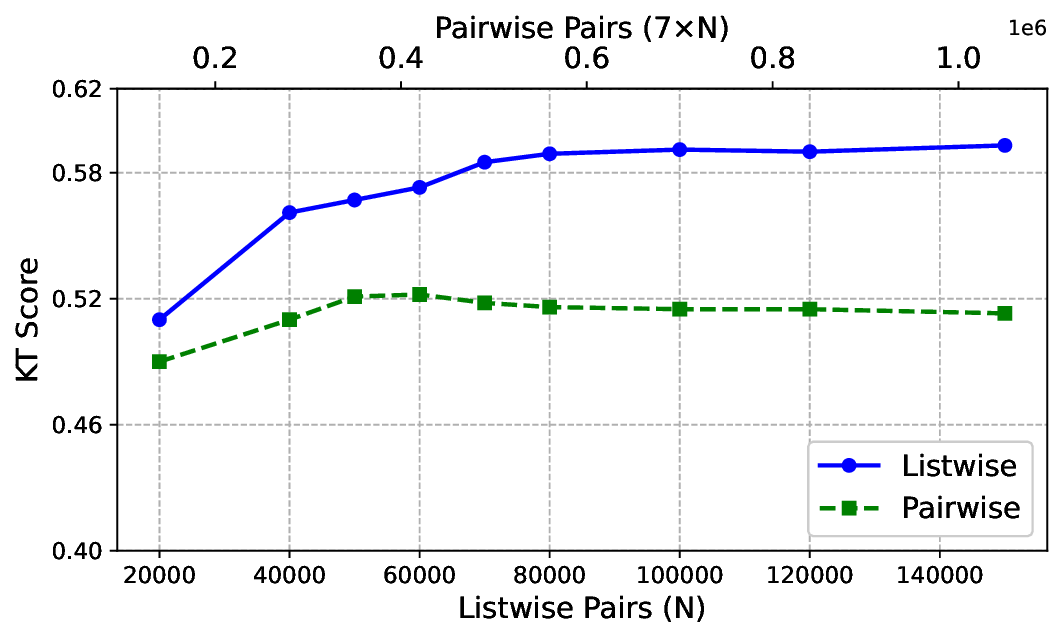}
		\label{fig:valence_test}
	}
	\subfigure[Performance versus margin]
	{
		\includegraphics[width=0.75\columnwidth]{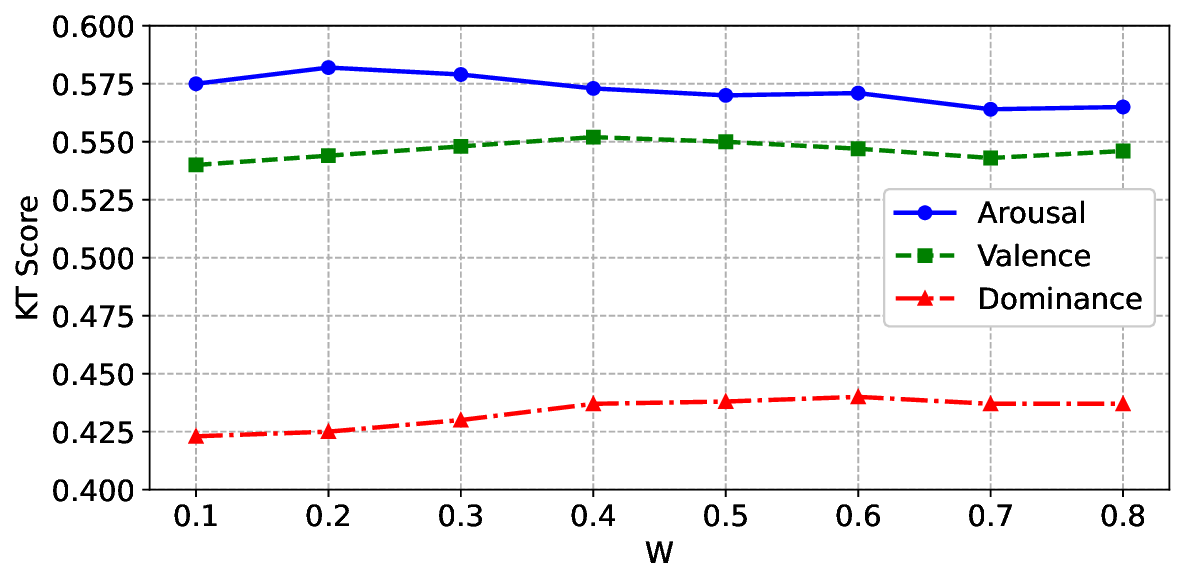}
		\label{fig:arousal_test}
	}
	\caption{(a) Comparison of Pairwise vs. Listwise training with increasing number of visible comparisons (arousal attribute only). (b) Impact of minimum margin $W$ on RankList performance across arousal, valence, and dominance.}
	\label{fig:ablation_additional}
\end{figure}

To better understand the behavior of our RankList framework, we conduct two ablation studies on training configuration and sampling strategies. Figure~\ref{fig:ablation_additional}(a) shows that RankList outperforms pairwise RankNet consistently as training size increases. For fairness, we match the number of visible comparisons by using $7N$ pairwise pairs per $N$ listwise samples. While both methods improve with more data, RankList maintains steady gains, whereas pairwise learning saturates or slightly degrades, highlighting the superior generalization of listwise learning. Figure~\ref{fig:ablation_additional}(b) examines the effect of the minimum margin $W$ used during list construction (Eq.~\ref{eq:margin}). We vary $W$ from 0.1 to 0.8 and find that moderate values (0.3–0.5) offer the best performance. Dominance shows higher sensitivity to $W$, indicating that larger margins may be needed to model this attribute effectively.


\begin{table*}[t]
\centering
\fontsize{8}{11}\selectfont
\caption{Cross-corpus and within-corpus results using Kendall's Tau (KT) for arousal (A), valence (V), and dominance (D) across BIIC Podcast \cite{Upadhyay_2023_2}, IEMOCAP \cite{Busso_2008_5}, and MSP-IMPROV \cite{Busso_2017} datasets.}
\begin{tabular}{c|l|ccc|ccc|ccc}
\hline
\multirow{12}{*}[2.5ex]{\rotatebox[origin=c]{90}{\textbf{Cross-Corpus}}}
& \textbf{Method} & \multicolumn{3}{c|}{\textbf{BIIC Podcast}} & \multicolumn{3}{c|}{\textbf{IEMOCAP}} & \multicolumn{3}{c}{\textbf{MSP-IMPROV}} \\
& & A & V & D & A & V & D & A & V & D \\
\cline{1-11}
& Pairwise \cite{Burges_2005}       & 0.431 & 0.318 & 0.291 & 0.494 & 0.386 & 0.341 & 0.541 & 0.558 & 0.429 \\
& ListNet \cite{cao2007learning}    & 0.420 & 0.302 & 0.274 & 0.488 & 0.374 & 0.328 & 0.520 & 0.547 & 0.421 \\
& ListMLE \cite{xia2008listwise}    & 0.438 & 0.316 & 0.295 & 0.491 & 0.391 & 0.332 & 0.536 & 0.552 & 0.434 \\
& SoftRank \cite{Taylor_2008}       & 0.422 & 0.311 & 0.290 & 0.486 & 0.388 & 0.340 & 0.532 & 0.543 & 0.416 \\
& RnC \cite{zha2023rank}            & 0.426 & 0.308 & 0.288 & 0.492 & 0.384 & 0.344 & 0.531 & 0.540  & 0.432 \\
& \textbf{RankList}                 & 0.447$^{\ast}$ & 0.326 & 0.302$^{\ast}$ & 0.508$^{\ast}$ & 0.397$^{\ast}$ & 0.353 & 0.549 & 0.569$^{\ast}$ & 0.444$^{\ast}$ \\
\hline
\multirow{6}{*}{\rotatebox[origin=c]{90}{\textbf{Within-Corpus}}}
& Pairwise \cite{Burges_2005}       & 0.453 & 0.334 & 0.306 & 0.511 & 0.413 & 0.360 & 0.558 & 0.573 & 0.452 \\
& ListNet \cite{cao2007learning}    & 0.436 & 0.326 & 0.291 & 0.496 & 0.398 & 0.341 & 0.532 & 0.561 & 0.437 \\
& ListMLE \cite{xia2008listwise}    & 0.451 & 0.331 & 0.304 & 0.502 & 0.405 & 0.355 & 0.540 & 0.564 & 0.442 \\
& SoftRank \cite{Taylor_2008}       & 0.445 & 0.331 & 0.301 & 0.493 & 0.410 & 0.347 & 0.536 & 0.553 & 0.439 \\
& RnC \cite{zha2023rank}            & 0.472 & 0.343 & 0.315 & 0.514 & 0.422 & 0.370 & 0.556 & 0.579 & 0.461 \\
& \textbf{RankList}                 & 0.488$^{\ast}$ & 0.357$^{\ast}$ & 0.338$^{\ast}$ & 0.529$^{\ast}$ & 0.436$^{\ast}$ & 0.372 & 0.571$^{\ast}$ & 0.592$^{\ast}$ & 0.484$^{\ast}$ \\
\hline
\end{tabular}
\vspace{0.5mm}
\parbox{0.95\linewidth}{\footnotesize $^{\ast}$ indicates that the proposed RankList approach is significantly better than all other baselines.}
\label{tab:cross_corpus_results}
\end{table*}

\subsection{Generalization of RankList}
Table \ref{tab:cross_corpus_results} presents both cross-corpus and within-corpus evaluation results for the proposed RankList framework against several preference learning baselines on the BIIC Podcast, IEMOCAP, and MSP-IMPROV datasets. In the cross-corpus setting, the models are trained on the MSP-Podcast corpus and tested on unseen corpora. RankList achieves the highest KT scores in nearly all conditions. Interestingly, the traditional pairwise RankNet model exhibits the strongest cross-domain performance among the baselines, suggesting its robustness in scenarios with domain shift. However, RankList consistently surpasses all baselines, including RankNet, with relative gains of 3-6\% on average across attributes and datasets. This result underscores the advantage of modeling structured listwise relationships. In the within-corpus evaluations, where each model is trained and tested on the same dataset, RankList again achieves the best overall performance across emotional attributes. While the \emph{Rank-n-Contrast} (RnC) model stands out as the strongest baseline in these matched settings, RankList maintains an edge, particularly in capturing global ordinal structures. These trends affirm that while RnC leverages neighborhood consistency effectively, it lacks explicit list-level optimization. One exception to this trend is the dominance prediction task on the IEMOCAP, where RankList's performance is comparable to that of the RnC method. Nevertheless, RankList remains competitive and robust, highlighting its applicability in both matched and mismatched evaluation scenarios.

\section{Applications: Aesthetic Image Ranking}

We also validate our listwise preference framework on aesthetic image ranking, an inherently subjective task where absolute scores often fail to capture human visual preferences \cite{talebi2018nima, kong2016photo}. Datasets such as AVA \cite{murray2012ava} and the artistic image aesthetics assessment corpus \cite{jin2022towards} have enabled research in this area by providing large-scale relative annotations. This domain highlights the need for listwise preference learning formulations that go beyond fixed metrics, capture both local and global item interactions, and remain stable during training.

We rely on the artistic image aesthetics assessment dataset \cite{jin2022towards}, which is a large-scale collection of images annotated with aesthetic ratings aggregated from multiple human raters, making it well-suited for preference-based modeling. We use image features extracted from a ResNet-50 backbone pre-trained on ImageNet, followed by a fine-tuned scoring network trained using the listwise loss. During training, images are grouped into lists of size $N$, and ordered based on their mean aesthetic scores. We enforce a score margin to ensure clear relative preferences within each list (Eq. \ref{eq:margin}). Similar to the protocol followed in the \emph{style-specific art assessment network} (SAAN) \cite{jin2022towards}, we randomly split the 60,337 images from the dataset into 53,937 for train and 6,400 for test sets. 
The test set is used as a held-out  set to evaluate the model's ranking ability. We measure performance with the \emph{Spearman rank correlation coefficient} (SRCC) and KT, which are often used in aesthetic prediction.

\begin{table}[t]
    \centering
    \fontsize{8}{11}\selectfont
    \caption{Performance comparison on the Artistic Image Aesthetics dataset using Spearman's rank correlation coefficient (SRCC) and Kendall's Tau (KT).}
    \begin{tabular}{l|c|c}
        \hline
        \textbf{Method} & \textbf{SRCC} & \textbf{KT} \\
        \hline
        Pairwise \cite{Burges_2005} & 0.464 & 0.318 \\
        ListNet \cite{cao2007learning} & 0.443 & 0.304 \\
        ListMLE \cite{xia2008listwise} & 0.461 & 0.320 \\
        SoftRank \cite{Taylor_2008} & 0.440 & 0.297 \\
        SAAN \cite{jin2022towards} & 0.471 & 0.324 \\
        RnC \cite{zha2023rank} & 0.465 & 0.321 \\
        \textbf{RankList} & 0.493$^{\ast}$ & 0.332$^{\ast}$ \\
        \hline
    \end{tabular}
    \vspace{0.5mm}
\parbox{0.95\linewidth}{\footnotesize $^{\ast}$ indicates that the proposed RankList approach is significantly better than all other baselines.}
    \label{tab:image_ranking_results}
\end{table}

Table \ref{tab:image_ranking_results} summarizes the performance of our approach against standard baselines. RankList achieves the highest scores among all methods, with an SRCC of 0.493 and a KT of 0.332, indicating strong alignment with human-annotated aesthetic preferences. Among the baselines, ListMLE and SAAN show competitive performance, with SAAN slightly outperforming others due to its joint modeling of semantic and aesthetic cues. RnC also provides a strong contrastive baseline but falls short of RankList, which benefits from explicit listwise supervision. Traditional listwise models such as ListNet and SoftRank underperform in this setting, likely due to their limited capacity to model structured ranking dependencies. These results highlight the generalization of RankList to perceptual ranking tasks, demonstrating that modeling structured list-level constraints improves robustness in domains with subtle and subjective judgments like image aesthetics.

\section{Discussion and Conclusions}

The RankList framework extends the classical RankNet formulation into a robust listwise paradigm, effectively modeling structured ranking relationships for subjective tasks. By incorporating both adjacent and non-adjacent (skip-connected) pairwise constraints within ordered sequences, RankList captures richer supervision compared to isolated pairwise comparisons. The introduced log-sum-exp approximation significantly improves optimization stability by smoothing the gradient landscape, leading to better convergence relative to methods aggregating individual pairwise losses. Our empirical evaluations across the diverse tasks of speech emotion recognition and aesthetic image assessment consistently demonstrate that RankList outperforms strong baselines such as ListMLE, SoftRank, and contrastive methods such as RnC. This advantage is particularly notable in cross-domain scenarios, underscoring the generalization capabilities of RankList.

Nevertheless, RankList presents certain limitations. The log-sum-exp formulation aggregates multiple comparisons into one expression, potentially obscuring individual ranking errors, especially when differences in scores are subtle. While this formulation enhances efficiency in serial computation, it may be less optimal for parallel execution compared to the original loss composed of independent log terms. Additionally, the current implementation of skip-term modeling employs fixed skip distances without adaptive or task-specific weighting, unlike metric-driven methods such as LambdaRank \cite{burges_2006}. Furthermore, RankList assumes meaningful ordering of training examples, which might restrict its applicability in domains characterized by ambiguous or noisy preference signals. Overall, RankList provides a flexible and principled listwise preference learning framework that effectively leverages structured ranking information. Ablation analyses validate the individual contributions of its components, reinforcing its suitability for complex subjective ranking tasks.

\bibliography{aaai2026,reference}

\begin{thebibliography}{39}
\providecommand{\natexlab}[1]{#1}

\bibitem[{Burges, Ragno, and Le(2006)}]{burges_2006}
Burges, C.; Ragno, R.; and Le, Q. 2006.
\newblock Learning to rank with nonsmooth cost functions.
\newblock \emph{Advances in neural information processing systems}, 19.

\bibitem[{Burges et~al.(2005)Burges, Shaked, Renshaw, Lazier, Deeds, Hamilton,
  and Hullender}]{Burges_2005}
Burges, C.; Shaked, T.; Renshaw, E.; Lazier, A.; Deeds, M.; Hamilton, N.; and
  Hullender, G. 2005.
\newblock Learning to rank using gradient descent.
\newblock In \emph{International conference on Machine learning (ICML 2005)},
  89--96. Bonn, Germany.

\bibitem[{Burges(2010)}]{burges2010ranknet}
Burges, C.~J. 2010.
\newblock From ranknet to lambdarank to lambdamart: An overview.
\newblock \emph{Learning}, 11(23-581): 81.

\bibitem[{Busso et~al.(2008)Busso, Bulut, Lee, Kazemzadeh, Mower, Kim, Chang,
  Lee, and Narayanan}]{Busso_2008_5}
Busso, C.; Bulut, M.; Lee, C.; Kazemzadeh, A.; Mower, E.; Kim, S.; Chang, J.;
  Lee, S.; and Narayanan, S. 2008.
\newblock {IEMOCAP}: Interactive emotional dyadic motion capture database.
\newblock \emph{Journal of Language Resources and Evaluation}, 42(4): 335--359.

\bibitem[{Busso and Narayanan(2008)}]{Busso_2008}
Busso, C.; and Narayanan, S. 2008.
\newblock Recording audio-visual emotional databases from actors: a closer
  look.
\newblock In \emph{Second International Workshop on Emotion: Corpora for
  Research on Emotion and Affect, International conference on Language
  Resources and Evaluation (LREC 2008)}, 17--22. Marrakech, Morocco.

\bibitem[{Busso et~al.(2017)Busso, Parthasarathy, Burmania, AbdelWahab,
  Sadoughi, and {Mower Provost}}]{Busso_2017}
Busso, C.; Parthasarathy, S.; Burmania, A.; AbdelWahab, M.; Sadoughi, N.; and
  {Mower Provost}, E. 2017.
\newblock {MSP-IMPROV}: An Acted Corpus of Dyadic Interactions to Study Emotion
  Perception.
\newblock \emph{IEEE Transactions on Affective Computing}, 8(1): 67--80.

\bibitem[{Cao, Verma, and Nenkova(2015)}]{Cao_2015}
Cao, H.; Verma, R.; and Nenkova, A. 2015.
\newblock Speaker-sensitive emotion recognition via ranking: Studies on acted
  and spontaneous speech.
\newblock \emph{Computer Speech \& Language}, 29(1): 186--202.

\bibitem[{Cao et~al.(2007)Cao, Qin, Liu, Tsai, and Li}]{cao2007learning}
Cao, Z.; Qin, T.; Liu, T.-Y.; Tsai, M.-F.; and Li, H. 2007.
\newblock Learning to rank: from pairwise approach to listwise approach.
\newblock In \emph{Proceedings of the 24th international conference on Machine
  learning}, 129--136.

\bibitem[{Herbrich, Graepel, and Obermayer(2000)}]{herbrich2000large}
Herbrich, R.; Graepel, T.; and Obermayer, K. 2000.
\newblock Large margin rank boundaries for ordinal regression.
\newblock Technical report, Technical report, Microsoft Research.

\bibitem[{Jin et~al.(2022)Jin, Zhang, Li, Wang, and Wu}]{jin2022towards}
Jin, X.; Zhang, Y.; Li, Z.; Wang, M.; and Wu, X. 2022.
\newblock Towards artistic image aesthetics assessment: A large-scale dataset
  and a new method.
\newblock \emph{IEEE Transactions on Image Processing}, 31: 284--297.

\bibitem[{Joachims(2002)}]{joachims2002optimizing}
Joachims, T. 2002.
\newblock Optimizing search engines using clickthrough data.
\newblock In \emph{Proceedings of the eighth ACM SIGKDD international
  conference on Knowledge discovery and data mining}, 133--142.

\bibitem[{Kong and Shen(2016)}]{kong2016photo}
Kong, S.; and Shen, X. 2016.
\newblock Photo aesthetics ranking network with attributes and content
  adaptation.
\newblock In \emph{European Conference on Computer Vision (ECCV)}, 662--679.
  Springer.

\bibitem[{Lei and Cao(2023)}]{Lei_2023}
Lei, Y.; and Cao, H. 2023.
\newblock Audio-Visual Emotion Recognition With Preference Learning Based on
  Intended and Multi-Modal Perceived Labels.
\newblock \emph{IEEE Transactions on Affective Computing}, 14(4): 2954--2969.

\bibitem[{Li, Lin, and Li(2007)}]{li2007mcrank}
Li, P.; Lin, Q.; and Li, H. 2007.
\newblock Mcrank: Learning to rank using multiple classification and gradient
  boosting.
\newblock In \emph{Advances in neural information processing systems},
  897--904.

\bibitem[{Liu, Li, and Lee(2021)}]{Liu_2021_3}
Liu, A.~T.; Li, S.-W.; and Lee, H.-Y. 2021.
\newblock {TERA}: Self-Supervised Learning of Transformer Encoder
  Representation for Speech.
\newblock \emph{IEEE/ACM Transactions on Audio, Speech, and Language
  Processing}, 29: 2351--2366.

\bibitem[{Liu et~al.(2009)}]{liu2009learning}
Liu, T.-Y.; et~al. 2009.
\newblock Learning to rank for information retrieval.
\newblock \emph{Foundations and Trends{\textregistered} in Information
  Retrieval}, 3(3): 225--331.

\bibitem[{Lotfian and Busso(2016{\natexlab{a}})}]{Lotfian_2016}
Lotfian, R.; and Busso, C. 2016{\natexlab{a}}.
\newblock Practical considerations on the use of preference learning for
  ranking emotional speech.
\newblock In \emph{IEEE International Conference on Acoustics, Speech and
  Signal Processing (ICASSP 2016)}, 5205--5209. Shanghai, China.

\bibitem[{Lotfian and Busso(2016{\natexlab{b}})}]{Lotfian_2016_2}
Lotfian, R.; and Busso, C. 2016{\natexlab{b}}.
\newblock Retrieving Categorical Emotions using a Probabilistic Framework to
  Define Preference Learning Samples.
\newblock In \emph{Interspeech 2016}, 490--494. San Francisco, CA, USA.

\bibitem[{Lotfian and Busso(2019)}]{Lotfian_2019_3}
Lotfian, R.; and Busso, C. 2019.
\newblock Building Naturalistic Emotionally Balanced Speech Corpus by
  Retrieving Emotional Speech From Existing Podcast Recordings.
\newblock \emph{IEEE Transactions on Affective Computing}, 10(4): 471--483.

\bibitem[{Mote, Sisman, and Busso(2024)}]{Mote_2024}
Mote, P.; Sisman, B.; and Busso, C. 2024.
\newblock Unsupervised Domain Adaptation for Speech Emotion Recognition using
  {K-Nearest} Neighbors Voice Conversion.
\newblock In \emph{Interspeech 2024}, 1045--1049. Kos Island, Greece.

\bibitem[{Murray, Marchesotti, and Perronnin(2012)}]{murray2012ava}
Murray, N.; Marchesotti, L.; and Perronnin, F. 2012.
\newblock AVA: A large-scale database for aesthetic visual analysis.
\newblock \emph{IEEE Conference on Computer Vision and Pattern Recognition
  (CVPR)}, 2408--2415.

\bibitem[{Naini et~al.(2024)Naini, Kohler, Richerson, Robinson, and
  Busso}]{Naini_2024}
Naini, A.~R.; Kohler, M.; Richerson, E.; Robinson, D.; and Busso, C. 2024.
\newblock Generalization of self-supervised learning-based representations for
  cross-domain speech emotion recognition.
\newblock In \emph{IEEE International Conference on Acoustics, Speech, and
  Signal Processing (ICASSP 2024)}, volume To appear. Seoul, Republic of Korea.

\bibitem[{Naini, Kohler, and Busso(2023)}]{Naini_2023}
Naini, A.~R.; Kohler, M.~A.; and Busso, C. 2023.
\newblock Unsupervised Domain Adaptation for Preference Learning Based Speech
  Emotion Recognition.
\newblock In \emph{ICASSP 2023 - 2023 IEEE International Conference on
  Acoustics, Speech and Signal Processing (ICASSP)}, 1--5.

\bibitem[{Naini, Salman, and Busso(2023)}]{Naini_2023_2}
Naini, A.~R.; Salman, A.; and Busso, C. 2023.
\newblock Preference learning labels by anchoring on consecutive annotations.
\newblock In \emph{Interspeech 2023}, 1898--1902. Dublin, Ireland.

\bibitem[{Naini et~al.(2023)Naini, Subramanium, Leem, and Busso}]{Naini_2023_3}
Naini, A.~R.; Subramanium, S.; Leem, S.-G.; and Busso, C. 2023.
\newblock Combining relative and absolute learning formulations to predict
  emotional attributes from speech.
\newblock In \emph{IEEE Automatic Speech Recognition and Understanding Workshop
  (ASRU 2023)}, 1--8. Taipei, Taiwan.

\bibitem[{Parthasarathy, Lotfian, and Busso(2017)}]{Parthasarathy_2017}
Parthasarathy, S.; Lotfian, R.; and Busso, C. 2017.
\newblock Ranking Emotional attributes with Deep Neural Networks.
\newblock In \emph{IEEE International Conference on Acoustics, Speech and
  Signal Processing (ICASSP 2017)}, 4995--4999. New Orleans, LA, USA.

\bibitem[{Qin, Liu, and Li(2010)}]{qin2010general}
Qin, T.; Liu, T.-Y.; and Li, H. 2010.
\newblock A general approximation framework for direct optimization of
  information retrieval measures.
\newblock \emph{Information retrieval}, 13: 375--397.

\bibitem[{Qin et~al.(2008)Qin, Liu, Zhang, Wang, and Li}]{qin2008listwise}
Qin, T.; Liu, T.-Y.; Zhang, X.-D.; Wang, D.-S.; and Li, H. 2008.
\newblock A general approximation framework for direct optimization of
  information retrieval measures.
\newblock \emph{Information Processing \& Management}, 44(2): 477--491.

\bibitem[{Reddi et~al.(2016)Reddi, Hefny, Sra, Poczos, and
  Smola}]{reddi2016stochastic}
Reddi, S.~J.; Hefny, A.; Sra, S.; Poczos, B.; and Smola, A. 2016.
\newblock Stochastic variance reduction for nonconvex optimization.
\newblock In \emph{International conference on machine learning}, 314--323.
  PMLR.

\bibitem[{Swezey et~al.(2021)Swezey, Grover, Charron, and
  Ermon}]{swezey2021pirank}
Swezey, R.; Grover, A.; Charron, B.; and Ermon, S. 2021.
\newblock Pirank: Scalable learning to rank via differentiable sorting.
\newblock \emph{Advances in Neural Information Processing Systems}, 34:
  21644--21654.

\bibitem[{Talebi and Milanfar(2018)}]{talebi2018nima}
Talebi, H.; and Milanfar, P. 2018.
\newblock NIMA: Neural image assessment.
\newblock In \emph{IEEE Conference on Computer Vision and Pattern Recognition
  (CVPR)}, 586--594.

\bibitem[{Taylor et~al.(2008)Taylor, Guiver, Robertson, and
  Minka}]{Taylor_2008}
Taylor, M.; Guiver, J.; Robertson, S.; and Minka, T. 2008.
\newblock Softrank: optimizing non-smooth rank metrics.
\newblock In \emph{Proceedings of the 2008 International Conference on Web
  Search and Data Mining}, 77--86.

\bibitem[{Upadhyay et~al.(2023)Upadhyay, Chien, Su, Goncalves, Wu, Salman,
  Busso, and Lee}]{Upadhyay_2023_2}
Upadhyay, S.; Chien, W.-S.; Su, B.-H.; Goncalves, L.; Wu, Y.-T.; Salman, A.;
  Busso, C.; and Lee, C.-C. 2023.
\newblock An Intelligent Infrastructure Toward Large Scale Naturalistic
  Affective Speech Corpora Collection.
\newblock In \emph{International Conference on Affective Computing and
  Intelligent Interaction (ACII 2023)}. Cambridge, MA, USA.

\bibitem[{Wagner et~al.(2023)}]{Wagner_2023}
Wagner, J.; et~al. 2023.
\newblock Dawn of the Transformer Era in Speech Emotion Recognition: Closing
  the Valence Gap.
\newblock \emph{IEEE Transactions on Pattern Analysis \& Machine Intelligence},
  45(09): 10745--10759.

\bibitem[{Wu et~al.(2010)Wu, Burges, Svore, and Gao}]{wu2010adapting}
Wu, Q.; Burges, C.~J.; Svore, K.~M.; and Gao, J. 2010.
\newblock Adapting boosting for information retrieval measures.
\newblock \emph{Information retrieval}, 13(3): 254--270.

\bibitem[{Xia et~al.(2008)Xia, Liu, Wang, Zhang, and Li}]{xia2008listwise}
Xia, F.; Liu, T.-Y.; Wang, J.; Zhang, W.; and Li, H. 2008.
\newblock Listwise approach to learning to rank: theory and algorithm.
\newblock In \emph{Proceedings of the 25th international conference on Machine
  learning}, 1192--1199.

\bibitem[{Yannakakis, Cowie, and Busso(2017)}]{Yannakakis_2017}
Yannakakis, G.; Cowie, R.; and Busso, C. 2017.
\newblock The Ordinal Nature of Emotions.
\newblock In \emph{International Conference on Affective Computing and
  Intelligent Interaction (ACII 2017)}, 248--255. San Antonio, TX, USA.

\bibitem[{Yannakakis, Cowie, and Busso(2021)}]{Yannakakis_2021}
Yannakakis, G.; Cowie, R.; and Busso, C. 2021.
\newblock The Ordinal Nature of Emotions: An Emerging Approach.
\newblock \emph{IEEE Transactions on Affective Computing}, 12(1): 16--35.

\bibitem[{Zha et~al.(2023)Zha, Cao, Son, Yang, and Katabi}]{zha2023rank}
Zha, K.; Cao, P.; Son, J.; Yang, Y.; and Katabi, D. 2023.
\newblock Rank-n-contrast: Learning continuous representations for regression.
\newblock \emph{Advances in Neural Information Processing Systems}, 36:
  17882--17903.

\end{thebibliography}

\clearpage

\appendix

\section*{Appendix A: Discussion of the Approximation Terms}
\label{appendix:approx_terms}

We provide a detailed expansion of the approximation performed in transitioning from \( \mathcal{L}_{\text{extended}} \) to \( \mathcal{L}_{\text{approx}} \). The original extended loss is expressed as a product:

\begin{equation}
\mathcal{L}_{\text{extended}} = \log\left[\prod_{k=0}^{K}\prod_{i=1}^{N-k-1}(1 + e^{-\sigma O_{i(i+k+1)}})\right].
\end{equation}

Expanding the product inside the logarithm yields a combinatorial expansion of additive terms:
\begin{equation}
\begin{aligned}
\mathcal{L}_{\text{extended}} = \log\Big[1 
+ \sum_{k=0}^{K}\sum_{i=1}^{N-k-1} e^{-\sigma O_{i(i+k+1)}} \\
+ \sum_{\substack{(i,j)\\(k_1,k_2)}} e^{-\sigma(O_{i(i+k_1+1)} + O_{j(j+k_2+1)})} \\
+ \sum_{\substack{(i,j,l)\\(k_1,k_2,k_3)}} 
  e^{-\sigma (O_{i(i+k_1+1)} + O_{j(j+k_2+1)} + O_{l(l+k_3+1)})} \\
+ \dots \Big]
\end{aligned}
\end{equation}

This series includes single, double, triple, and higher-order exponential terms. Our approximation in \( \mathcal{L}_{\text{approx}} \) retains only the first-order additive terms, effectively neglecting second-order and higher-order interactions. This approximation is valid under the condition that pairwise differences \( O_{i(i+k+1)} \) are sufficiently large, making higher-order terms exponentially smaller in magnitude compared to first-order terms. Under typical practical training conditions, this assumption holds true as misordering margins are gradually increased during training. Neglecting these higher-order terms significantly simplifies gradient computations, ensures numerical stability, and prioritizes the most impactful comparisons during optimization. Thus, the approximation strikes a balance between computational tractability and fidelity to the original product formulation.

A natural question arises: if the simplified approximation removes terms from the original loss, why does it lead not only to faster training but also to improved performance? While truncating higher-order interactions reduces the numerical value of the loss by design, empirical analysis reveals that the gradient norm remains largely preserved, retaining over 91\% of its magnitude compared to the full extended loss. Specifically, we compared the accumulated loss and gradient magnitudes of the full extended formulation and our simplified approximation over a training epoch following RankNet pretraining. The accumulated loss under the approximation was approximately 0.78 times that of the extended loss, as expected due to the elimination of higher-order terms. However, the total gradient magnitude was about 0.91 times that of the full extended loss gradient. This observation indicates that our approximation preserves the core optimization signal while filtering out redundant or low-impact components. This behavior is consistent with insights from optimization theory, where simplified or variance-reduced gradient formulations have been shown to promote stable and effective convergence in non-convex settings \cite{reddi2016stochastic}. By focusing gradient contributions on dominant pairwise terms, the model reduces the influence of less prominent interactions, allowing the optimization to emphasize meaningful rank order corrections. Consequently, the simplified loss not only improves computational efficiency but also enhances the performance on perceptual ranking tasks such as speech emotion recognition and aesthetic assessment.

A key reason the RankList approximation remains robust, even after removing higher-order terms, is due to how noise affects the additive versus multiplicative structure of the loss. In RankList, the simplified loss takes the form:

\begin{equation}
\mathcal{L}_{\text{RankList}} = \log\left(1 + \sum e^{-\sigma O_{ij}}\right),
\end{equation}

\noindent
where each term \( e^{-\sigma O_{ij}} \) is added inside the logarithm. If one pair \( (i,j) \) is incorrectly ordered, its influence enters additively and is diluted among many other small-magnitude terms.

In contrast, the full extended loss is:

\begin{equation}
\mathcal{L}_{\text{Extended}} = \log\left(\prod (1 + e^{-\sigma O_{ij}})\right),
\end{equation}

which multiplies all pairwise contributions before taking the logarithm. In this formulation, a single erroneous term (where \( O_{ij} < 0 \)) can disproportionately skew the product, thus amplifying its influence on both the loss and its gradient. Therefore, RankList not only simplifies computation but also inherently reduces the impact of spurious or noisy pairwise terms by modeling them within a summation rather than a product.

\section*{Appendix B: Baseline Methods}

This section provides an expanded description of all baseline methods used for comparison in our experiments. Each method is explained with its key formulation, optimization objective, and conceptual difference from our proposed RankList framework.

\subsection*{B.1 Pairwise RankNet \cite{Burges_2005}}
RankNet is a pairwise preference learning model that predicts the relative ordering between pairs of samples. Given two samples $x_i$ and $x_j$, RankNet models the probability that $x_i$ should be ranked higher than $x_j$ based on their scores $s_i = f(x_i)$ and $s_j = f(x_j)$. The probability is computed using a logistic function:

\begin{equation}
P_{ij} = \frac{1}{1 + \exp(-\sigma(s_i - s_j))},
\end{equation}

where $\sigma$ is a scaling parameter. The ground-truth preference $\bar{P}_{ij}$ is typically binary:

\begin{equation}
\bar{P}_{ij} = \begin{cases}
1 & \text{if } x_i \text{ is preferred over } x_j, \\
0 & \text{otherwise.}
\end{cases}
\end{equation}

The loss is the cross-entropy between predicted and target preferences:

\begin{equation}
\mathcal{L}_{\text{RankNet}} = -\bar{P}_{ij} \log P_{ij} - (1 - \bar{P}_{ij}) \log(1 - P_{ij}).
\end{equation}

Unlike listwise methods, RankNet considers each pair in isolation and does not model the overall list structure. Our RankList generalizes RankNet to structured lists, enabling modeling of both local and global constraints.

\subsection*{B.2 ListNet \cite{cao2007learning}}
ListNet introduces a listwise learning-to-rank formulation by modeling a probability distribution over permutations of items. Given a score vector $\mathbf{s} = [s_1, s_2, \dots, s_N]$, the permutation probability is defined as:

\begin{equation}
P(\pi | \mathbf{s}) = \prod_{i=1}^{N} \frac{\exp(s_{\pi(i)})}{\sum_{k=i}^{N} \exp(s_{\pi(k)})}.
\end{equation}

The loss function minimizes the cross-entropy between the predicted and ground-truth permutation distributions. Due to the factorial complexity of full permutations, ListNet typically uses a top-one probability approximation:

\begin{equation}
P_{\text{top-1}}(i | \mathbf{s}) = \frac{\exp(s_i)}{\sum_{j=1}^{N} \exp(s_j)}.
\end{equation}

Then, the loss becomes:

\begin{equation}
\mathcal{L}_{\text{ListNet}} = -\sum_{i=1}^{N} P_{\text{top-1}}(i | \pi^*) \log P_{\text{top-1}}(i | \mathbf{s}).
\end{equation}

While ListNet is listwise in nature, it relies heavily on approximated permutations and does not explicitly model pairwise misorderings or long-range list interactions.

\subsection*{B.3 ListMLE \cite{xia2008listwise}}
ListMLE formulates listwise ranking as a maximum likelihood estimation problem. It seeks to maximize the likelihood of observing the ground-truth permutation $\pi^*$ given the predicted scores $\mathbf{s}$. The probability of $\pi^*$ is:

\begin{equation}
P(\pi^* | \mathbf{s}) = \prod_{i=1}^{N} \frac{\exp(s_{\pi^*(i)})}{\sum_{k=i}^{N} \exp(s_{\pi^*(k)})}.
\end{equation}

The loss function is the negative log-likelihood:

\begin{equation}
\mathcal{L}_{\text{ListMLE}} = -\log P(\pi^* | \mathbf{s}).
\end{equation}

ListMLE focuses on modeling the entire permutation using the Plackett-Luce distribution. It captures more structure than pointwise or pairwise methods but still assumes noise-free full permutations, which are rare in subjective domains like emotion or aesthetics.

\subsection*{B.4 SoftRank \cite{Taylor_2008}}
SoftRank provides a differentiable approximation to rank-based metrics like NDCG by modeling the rank of each item as a random variable. It estimates the probability that one item ranks higher than another using the logistic function:

\begin{equation}
P(s_i < s_j) = \frac{1}{1 + \exp(s_i - s_j)}.
\end{equation}

The expected rank $\mu_i$ of each item is then computed as:

\begin{equation}
\mu_i = 1 + \sum_{j \neq i} P(s_j > s_i).
\end{equation}

A Gaussian smoothing is applied to convert expected ranks into a probability distribution over discrete rank positions. The loss is the cross-entropy between this predicted rank distribution and the ground-truth rank:

\begin{equation}
\mathcal{L}_{\text{SoftRank}} = -\sum_{i=1}^{N} \sum_{r=1}^{N} P(r_i = r | y_i) \log P(r_i = r | \mu_i).
\end{equation}

SoftRank provides gradient flow via smoothed ranks, but it may suffer from gradient vanishing or convergence issues due to soft averaging of many terms.

\subsection*{B.5 Rank-n-Contrast (RnC) \cite{zha2023rank}}
Rank-n-Contrast (RnC) is a contrastive framework designed to learn regression-aware representations in continuous-valued tasks. It constructs local neighborhoods based on scalar targets (e.g., arousal/valence scores) and defines positives and negatives based on relative proximity in the label space.

Given anchor sample $x_i$ and its embedding $z_i = f(x_i)$, the loss is:

\begin{equation}
\mathcal{L}_{\text{RnC}} = -\log \frac{\exp(\text{sim}(z_i, z_j^+)/\tau)}{\exp(\text{sim}(z_i, z_j^+)/\tau) + \sum_{j^-} \exp(\text{sim}(z_i, z_{j^-})/\tau)},
\end{equation}

\noindent
where $\text{sim}(\cdot, \cdot)$ is the cosine similarity and $\tau$ is a temperature. RnC encourages embeddings to preserve local relative orderings but does not explicitly model full list structure. Unlike RankList, RnC learns from unordered contrastive pairs and lacks list-level supervision.

\subsection*{B.6 SAAN \cite{jin2022towards}}
The \emph{semantic-aesthetic alignment network} (SAAN) is a regression model developed for aesthetic image assessment. It uses a two-stream architecture with semantic and aesthetic encoders (e.g., Swin Transformer backbones) to produce a quality score.

Given an image $x$ and its aesthetic score $y$, the model outputs $\hat{y} = f(x)$, and the training loss is:

\begin{equation}
\mathcal{L}_{\text{SAAN}} = (y - \hat{y})^2.
\end{equation}

Although SAAN achieves strong regression performance, it does not model relative or listwise preferences. It assumes pointwise supervision and focuses on minimizing absolute prediction errors. In contrast, RankList focuses on preserving ranking structure, which is more aligned with subjective tasks like aesthetics or emotion perception.

\section*{Appendix C: Dataset Details}

This section provides detailed descriptions of the speech emotion recognition datasets used in our experiments. All datasets provide continuous annotations for emotional attributes such as arousal, valence, and dominance, enabling the construction of preference-based rankings for training and evaluation.

\subsection*{C.1 MSP-Podcast Corpus}
We use release 1.12 of the MSP-Podcast corpus \cite{Lotfian_2019_3}, a large-scale emotional speech database comprising over 324 hours of audio. The recordings were collected from online platforms under Creative Commons licenses, spanning diverse conversational topics including science, politics, entertainment, and personal narratives. Strict quality filters were applied to remove segments containing background noise, overlapping speech, or music. Each segment is annotated by at least five raters using dimensional emotion attributes arousal, valence, and dominance which respectively capture the perceived intensity, polarity, and control conveyed by the speaker. Although categorical labels are also available, our experiments focus exclusively on these continuous emotional attributes. The dataset is partitioned into training (112,712 segments), development (31,961 segments), and test (44,395 segments) sets, ensuring speaker independence across partitions. All preference-based rankings were generated within these respective subsets.

\subsection*{C.2 IEMOCAP Corpus}
The USC-IEMOCAP corpus \cite{Busso_2008} is a widely-used multimodal emotion dataset consisting of approximately 12 hours of dyadic interactions between ten professional actors across five sessions. The dataset includes a total of 10,527 speaking turns featuring both scripted and spontaneous conversations. Emotional annotations are provided at the turn level, including both discrete emotion categories (e.g., happy, sad, angry) and dimensional ratings for arousal, valence, and dominance. These attributes are labeled on a discrete scale from 1 to 5 by multiple annotators. Given its balance of emotional breadth and annotation precision, IEMOCAP remains a standard benchmark in the emotion recognition community.

\subsection*{C.3 MSP-IMPROV Corpus}
The MSP-IMPROV corpus \cite{Busso_2017} contains dyadic interactions between 12 professional actors across six sessions, totaling 8,438 speaking turns. Unlike the MSP-Podcast corpus, which is sourced from naturalistic web audio, the MSP-IMPROV corpus was recorded in a controlled lab setting, offering clean audio conditions. Each segment is annotated by at least five annotators for emotional attributes, specifically arousal, valence, and dominance. This makes MSP-IMPROV an ideal benchmark for evaluating domain mismatch effects and testing generalization. In our experiments, we use data from the first three sessions (six speakers) as the test set.

\subsection*{C.4 BIIC-Podcast Corpus}
The BIIC-Podcast corpus \cite{Upadhyay_2023_2} contains approximately 157 hours of speech collected from Taiwanese Mandarin podcasts. Following the methodology used in the MSP-Podcast corpus, all utterances are annotated with arousal, valence, and dominance scores, along with primary and secondary categorical emotion labels. The audio content spans a broad range of everyday conversations and thematic topics. This dataset serves as a cross-lingual evaluation benchmark in our experiments, allowing us to test the robustness of preference learning models under significant domain and language shifts.

\end{document}